# An Object-Oriented and Fast Lexicon for Semantic Generation

*Maarten Hijzelendoorn and Crit Cremers*

Leiden University Centre for Linguistics (LUCL)

## 1. Introduction

Delilah consists of a parser and a generator for complex Dutch sentences, in (ISO) Prolog. It yields full syntactic and semantic representations by unification of comprehensive lexical structures, on the basis of a combinatory categorial grammar (Cremers 2002) and accompanying higher-order lambda terms. The lexical structures are HPSG-style, but Delilah is driven by multimodal combinatory categorial grammar (Cremers 1993, Moortgat 1997, Steedman 2000, Baldridge and Kruijff 2003). The main characteristic of this grammar is that it combines rigid syntactic structure with compositional semantics. To our knowledge, Delilah (http://www.delilah.eu) is actually the only operational semantic generator and parser for Dutch.

Delilah represents the 'gnostic' approach to computational linguistics: the view that explicit knowledge of language can be assembled, formalized and exploited up to full semantic interpretation. In this sense, it constitutes an alternative to predominantly statistical and sub-symbolic approaches to natural language processing. Since it focuses on semantics and the full specification of logical form, it must operate on a high level of grammatical and lexical fine-grainedness. As a consequence, Delilah's lexicon is both detailed and large. It is generated on the basis of a restricted number of underspecified, generic templates, like 'transitive verb', representing minimal default graphs for all kinds of lexical types. These templates can also be considered as constructions in the sense of Croft (2001). The templates formulate important generalizations about lexical relations, meanings and syntactic behavior. This class of templates is flexible, and defined by practical and empirical considerations. A particular lemma is produced by specifying differences with respect to a particular template. In this sense, every lemma itself defines a template. A set of rules produces the lemma from the generic template(s) by inheritance and the specified difference list. From this lemma, another set of rules produces graphs for all the morphologically or otherwise marked instances of the lemma. The lexicon, a set of 'lexical entries', thus hinges on three components: a set of generic templates ('lexical units'), a set of lemma difference specifications and an utterly complex set of derivational rules. Data management, then, is done on a higher level than in a traditional database. The lexicon is generated off-line, which is repeated when the data scheme, or one or more generic templates, lemmas, or rule sets have changed. Figure 1 is an example of a lemma.

```
|ID:A+B
|HEAD:|CONCEPT:discover
|     |PHON:ontdekt
|     |SLF:discover
|     |SYNSEM:|ETYPE:event
|     |       |FLEX:fin
|     |       |NUMBER:sing
|     |       |PERSON:or([2,3])
|     |       |TENSEOP:at-pres
|     |       |VTYPE:transacc
|PHON:C
|PHONDATA:lijnop(ontdekt,A+B,[arg(right(-1),0,D),arg(left(11),wh,E)],C)
|SLF:{{[F&(B+G)#H,I&(B+J)#K,L@some^M^and(quant(M,some),discover~[M],event~[M],entails1(M,
incr),and(L,entails(M,incr)))&(A+B)#N],[],[]],and(and(and(agent_of~[N,H],theme_of~[N,K]),
attime(N,O)),tense(N,pres))}
|SYNSEM:|CAT:s
|       |EVENTVAR:N
|       |EXTTH:agent_of~[A+B,H]
|       |PREDTYPE:nonerg
|       |SUBQMODE:P
|       |TENSE:tensed
|TYPE:s\0~[np^wh#B+G]/0~[np^0#B+J]
|ARG:|ID:B+G
|    |PHON:E
|    |SLF:F
|    |SYNSEM:|CASE:nom
|    |       |CAT:np
|    |       |NUMBER:sing
|    |       |OBJ:subject_of(A+B)
|    |       |PERSON:or([2,3])
|    |       |QMODE:P
|    |       |THETA:agent_of
|ARG:|ID:B+J
|    |PHON:D
|    |SLF:I
|    |SYNSEM:|CASE:obliq
|    |       |CAT:np
|    |       |OBJ:dirobject_of(A+B)
|    |       |THETA:theme_of
```

Figure 1: A lemma for 'ontdekt' *discovers* (2nd/3rd pers. pres. sing)

The lemmas are completely defined by HPSG-style feature-value specifications (Sag et al. 2003). Lemmas are complex symbols, and can be represented by Directed Acyclic Graphs (DAGs). Typically, they have a different number of features. A lemma may or may not specify a certain value for a certain feature. In the latter case, the lemma is underspecified. Besides atoms and numbers, values can be complex structures themselves, defining sub-graphs. A lemma will contain sub-graphs for semantically and/or syntactically related phrases. Unification will apply to these sub-graphs, that is, two graphs A and B unify whenever B unifies with a designated sub-graph of A, in which case A is called primary and B a secondary. By definition, the primary graph constrains the secondary graph in every relevant aspect: morphologically, syntactically and semantically. Thus, the Delilah lemma is a natural way of expressing collocational effects, from weak combinatory effects to rigid combinations. In fact, every lemma defines the domain for collocational effects. The lemma essentially separates constraints on sub-phrases of a structure from properties of the overall phrase. Inheritance and co-indexing are specified by using the same variable as value at different places in the graph.

The lexicon is a collection of explicitly defined, spelled-out linguistic entities. They are 'unrelated' in the sense that they are stored independently of each other, and 'autonomous' in the sense that they, once retrieved, are operated independently of each other. A different approach is followed in Cornetto, which defines a combinatorial and relational, i.e. implicit, network on word level (Vossen et al. 2007). In Delilah such an information network, including, for example, collocations, has been 'compiled away', yielding real linguistic entities to start with, e.g. for generation purposes.

The following illustrates the lexicon's size and growth and the storage and access problem of a large computational lexicon. Adding the lemma for *hij* 'he' means adding 1 lexical entry, while adding the lemma for *gelopen* 'walked' (past. part.) means adding 19 entries, adding the lemma for *heeft* 'has' (3rd.pers. pres. sing. aux) means adding 133 entries, and adding the whole paradigm for *verven* 'to paint' means adding 226 entries. Clearly, the fully written-out specification of lexical entries introduces an exponential storage factor. (These figures are to be interpreted relatively and don't mean anything on their own. They reflect the current state of the lexicon.) Furthermore, for Delilah's grammar-driven generation component efficient access to the lexicon is crucial, because a word form should only be produced when its lexical specification matches certain constraints specified by the grammar and by the generation algorithm. It has been observed that searching and finding of lexical entries is the main business of the generator. Therefore, efficient access methods are required for retrieval, with the property to search and match complex lexical graphs and lexical constraints, which is hard. Finally, we developed our system in Prolog (Clocksin and Mellish 1984) for historical reasons. We implemented standards as much as possible, including ISO Prolog (Deransart et al. 1996). Our problem, then, is implementing a fast and large-scale lexicon in a Prolog environment.

The rest of the paper is organized as follows. Section 2 evaluates two models for database management: the Relational Model and the Object-Oriented Model. Section 3 describes methods that are needed to build and access an object-oriented lexicon. Section 4 concludes the paper.

## 2. Models

We will concentrate on a computational model of a lexicon. Such a lexicon can be stored either in internal, working memory of the computer, or in external memory, e.g. on a hard disk. Storing it internally means at least loading all lexical entries, which takes much time when the lexicon is big. Access to data structures in internal memory is usually very fast, but Prolog's internal database is not so fast. Working memory cannot be extended beyond some gigabytes, which is not large enough for our purposes, while extension by virtual memory gives bad performance. The enormous proportions of the lexicon, let alone its foreseen expansion, and the limitations of current hardware rule out the option of storing it in internal memory. Storing it externally has no space problem, because hard disks are 'big enough' and cheap nowadays. An external medium, however, is slower, and harder to access. As the lexicon is of a static nature, once it has been generated, a full-featured database management system, including update facilities, is unnecessary. We can restrict ourselves to implementing a lexicon as a saved set of 'read-only' lexical entries, and provide efficient access methods for at least the most important features used by the generator, being the syntactic type, the semantic concept and the word form.

The lexicon, built as a collection of lexical entries, can be regarded as a set of records in a database. There are a number of database models for database management. The Relational Model (Codd 1970) is based on two parts of mathematics: first order predicate logic and the theory of relations. It is data-based and well-known from relational database management systems (RDBMS). As the programming language Prolog is based upon the Relational Model, it seems straightforward to consider this model in more detail. Alternatively, the Object-Oriented Model (Meyer 1997) is investigated, because it was noted that a lexical entry can be seen as a 'linguistic object' in the OO Model. Such an object stores data (lexical specification), and procedures (e.g. for linearization). The OO Model is knowledge-based. Furthermore, the OO Model is often recommended when there is a need for high performance processing on complex data, e.g. binary multimedia objects. The OO Model is well-known from graphical user interfaces, while object database management systems are arising.

## 2.1 The Relational Model

The Relational Model was the first formal database model, solidly founded on well-understood mathematical principles and explained by Date (2003). It was invented in those days when computer memory was scarce and expensive. A relational database consists of a number of relations ('tables'), in which all data is stored. Each relation is a set of tuples that all contain the same attributes (the horizontal rows, 'records'). A tuple is an unordered set of attribute values (the vertical columns, 'fields'). An $n$-tuple is an unordered set of $n$ values of $n$ attributes. All of the attribute values should be in the same domain, that is they should be a valid value for the data type of the attribute, and they should obey the same constraints. Data types must be scalar, like integer, or string, and cannot be compound, like a graph. Constraints provide a way of restricting the data that can be stored, either in tuples, or in attributes. A relation is said to be $n$-ary iff it consists of a set of $n$-tuples. A special kind of constraint is a key. A key is an $m$-tuple of an $n$-ary relation, where $m < n$, that enforces the uniqueness of the combination of the $m$ attribute values for each tuple. Key values are usually kept in an index or hash table, which is stored in internal memory for fast access. Keys prevent storing duplicate data. A relational variable can be assigned a subset of tuples as result of a query. Queries are stated in a query language, typically SQL (Chamberlin and Boyce 1974). A query can be simplex or complex, which implies consulting one or more tables, resp., possibly under the condition of some constraints, of which the key is the most important one, and which links tables.

### 2.1.1 The lexicon and the Relational Model

Our lexical entries are complex, non-atomic data structures. The Relational Model only allows atomic data types. It would be possible to pre-compile them into flat, atomic strings. However, flattening structured information is in general not a good idea, when it comes to retrieval on the basis of some highly detailed substructure.

Lexical entries are DAGs, that is recursive data structures. Although it would be possible to pre-compile all feature-value paths of a DAG to a number of tuples in different tables by means of a recursive procedure, it would be impossible to retrieve them, as a relational database system does not provide for recursive processing (Hirao 1990). On the other hand, when we regard the lexicon as knowledge, we can linearize and store the graphs into a relational database, and use the powerful processing of recursion for inferences by Prolog, Prolog being mobilized as a powerful query language. It is possible to successfully and easily store objects in a relational database by following a step-by-step procedure (Ambler 2000). A disadvantage is that quite a number of tables might be involved, as graphs typically hold large numbers of features, while for semantic generation no less than complete lexical specifications are required. Pre-compiling a lexical entry for a noun will typically yield another number of tuples (in just as many tables) as pre-compiling a verbal entry. The top-level, a main table, which is to represent complete lexical entries, has to span all the tuples into one large main tuple, in which each attribute represents a path, and where the attribute's value is either the value of the path, or an ID that links to another table. This implies that there will be more than one top-level: one table for each combination of attributes and consequently more than one main table. As we don't want to impose a restriction on the internal dependencies of a graph, there is no restriction on the recursion depth of features. Thus, we allow for dependencies in multi-word expressions of any kind. Consequently, the number of different main tables can be large in practice, which means decreasing performance with orders of magnitude. Thus, lexical entries cannot be regarded as single, homogeneous relations in the Relational Model, let alone be retrieved.

Furthermore, lexical entries use variables. The Relational Model does not allow attribute values to be variables. A variable is not an atomic constant, and thus not distinguishable from other values for the same attribute. As a consequence, an attribute that has a variable in its data domain cannot be indexed, which could lead to bad overall performance of the database. It would be possible to pre-compile variables into constants, and to de-compile them during retrieval. Calling meta-predicates, however, is in general rather time-consuming.

We conclude that the Relational Model cannot efficiently accommodate (logic) variables, recursive features and, consequently, the top-level. It does not meet high-performance demands on complex (recursive) data structures. It is inappropriate for our purposes.

## 2.2 The Object-Oriented Model

In the Object-Oriented Model, information and control is represented in the form of interacting 'objects', as is well-known from the Object-Oriented Programming (OOP) paradigm. An object can be seen as an information processor. It accepts commands from other objects, processes data by executing procedures ('methods'), which both are stored in the object, and sends commands to other objects to be executed by those objects. By keeping data (properties) and procedures (operations) together in one local unit, an object holds all characteristics related to some concept. This implies that an object is a complex data structure. It is independent of other objects, it has its own ID, and its own role. OOP, then, is modeling a problem by distinguishing different (abstract) levels of objects (called 'classes' and 'subclasses', which are maintained in a 'class hierarchy'), and defining their cooperation and interactions. A class defines the general characteristics of a concept in terms of the problem domain. An object is a particular instance of a class, from which it inherits all properties and methods, and to which it may add own information, or overwrite inherited information. Multiple inheritance is inheriting from more than one class, combining properties and methods. Classes are the structuring elements (modules) in OOP, which hide the details of the code to be accessed by objects that stem from other classes.

### 2.2.1 Prolog and the Object-Oriented Model

Prolog is a declarative programming language with different procedural semantics than the OOP paradigm. In Prolog the term 'object' refers to things that can be represented by terms, Prolog's only data type. It does not refer to a data structure that inherits from a class hierarchy. Some vendors have extended Prolog with object-oriented features (SICS 2008). However, in general, mixing programming styles will lead to unmanageable and incompatible code. Therefore, we will stick to Prolog's execution mechanism and to its term data type, and explore the possibility of objects as a data representation for lexical entries or: linguistic objects.

### 2.2.2 The lexicon and the Object-Oriented Model

Linguistic objects are generated by deriving information from one or more generic classes of templates, called 'constructions', and by adding local information. This fits nicely in the OOP concept of objects that are constructed by a specialized 'constructor' method and by (multiply) inheriting information from classes. A linguistic object, a complex, recursive graph, can be mapped onto an OOP object, which is a complex data structure. Shared variables are, in fact, an abbreviation for a unification procedure, deferred until runtime. Encoding them by an OOP method is straightforward. Linguistic objects are independent units and, thus, uniquely identifiable, as are OOP objects. This makes an object, including all properties and methods, accessible by one ID, which is utterly important for efficient storage and access by data-intensive processes, such as semantic generation. ISO Prolog terms, being complex data types, are well-equipped to represent OOP objects, including variables, recursion, and unique identification. On the other hand, linguistic objects are built from classes and any combinatory difference is compiled out as a difference between objects. These objects may differ from each other minimally, yielding a significant amount of overlap and introducing an exponential space factor. For example, the objects for the 2nd and 3rd person of a regular verb only differ in the person and phonological features. Objects can adopt different states when they get involved in some process. 'Persistent' objects are objects whose initial state have been saved.

We conclude that the Object-Oriented Model provides a natural environment for representing linguistic objects. It lifts the drawbacks of the Relational Model with respect to data modeling, and potentially enables fast access by unique identifiers. Its data redundancy is a small price to pay, given the considerable decrease of the price/performance ratio of hard disks each year. Future developments in runtime file (de-)compression techniques might weaken this disadvantage. Prolog's term data type is suitable to represent linguistic data objects.

## 2.3 Object-Oriented Databases

An Object-Oriented database system must satisfy two criteria: it should be a database management system, and it should be an object-oriented system, i.e., to the extent possible, it should be consistent with the current crop of object-oriented programming languages (Atkinson et al. 1989). OO databases arose when it was discovered that relational databases lacked high-performance processing on complex data structures. In the Relational Model complex data, being split and stored in several tables, have to be retrieved by searching these tables and combining ('joining') pieces of data, while in the Object-Oriented Model a complete object is retrieved by its ID in one operation. In the Relational Model the relational database is accessed by means of a declarative query language, typically SQL, which needs a procedural interpretation at runtime. This language permits general-purpose queries and transforms them into efficient retrieval procedures. In the Object-Oriented Model, lacking a standard query language, the OO database is accessed by means of a pre-compiled pointer mechanism, which can be regarded as an optimized query answer. In our application of a computational lexicon for semantic generation, we do need specialized queries, e.g. on syntactic type, semantic concept and word form. Hence, an OO database is appropriate.

We come to the conclusion that, as our lexicon is static after it has been derived from generic templates, we do not need a full-blown object database management system (ODBMS) for persistent storage, but we can stick to a simpler OO lexicon, listing fully specified linguistic objects, and for retrieval only. We can represent objects by Prolog's native term data type, and manage them by Prolog's execution mechanism and pre-compiled pointers.

## 2.4 An example

We illustrate the operation of the generator. It produces free, grammatical and meaningful sentences. In categorial grammar, a category consists of a head, and zero or more arguments to its left and/or right side. For generation purposes, the category can be seen as an agenda. The generation algorithm keeps already produced heads in an unordered list, and still to be produced arguments on a stack. It handles them by inserting and deleting elements into/from an arbitrary position, or onto/from the top of the stack position, respectively. It starts with a random semantic concept, and finds one of its realisations in the lexicon. The head of its category is inserted in the list. The arguments are shifted on the stack. Each argument is produced by either reduction with some head in the list, or by reduction with a new category to be found in the lexicon that has it as head. The topmost argument of the stack is replaced by the arguments of the new category or removed completely when it does not have arguments of itself. When the argument stack is empty, there are two possibilities. If the heads list does not consist of exactly one of the sentential categories s ('sentence') or q ('query'), the lexicon is consulted for a category that has the non-sentential category as an argument. Its head is inserted in the heads list, while its arguments are shifted on the argument stack. Otherwise, the algorithm stops. Categories, being complex symbols, are unified on each reduction step, yielding both linearization and underspecified logical form. In table 1 the procedure for deriving the sentence *die Nederlander ontdekt diepe betekenissen* 'that Dutchman discovers deep meanings' is listed. Only categories and number features are shown, instead of full complex symbols.

| Step | Action | Result | Heads | Args |
| --- | --- | --- | --- | --- |
| #1 | search for random concept | *betekenissen* 'meanings', cat=n, num=plur | {} | {} |
| #2 | insert head n, num=plur | | {$n_{plur}$} | - |
| #3 | search entry with arg n, unifiable with num=plur | *diepe* 'deep', cat=np/n, where n has num=plur | - | - |
| #4 | reduce arg n in #3 with head n in #1; insert head np, num=plur | *diepe betekenissen* 'deep meanings' | {$np_{plur}$} | - |
| #5 | search for entry with arg np, unifiable with num=plur | *ontdekt* 'discovers', cat=s\np1/np2, where s has num=sing, np1 has num=sing and np2 has num unspecified | - | - |
| #6 | reduce arg np2 in #5 with head np in #4; insert head s, num=sing; shift arg np1, dominated by s, num=sing | *ontdekt diepe betekenissen* 'discovers deep meanings'; np2 is assigned num=plur | {$s_{sing}$} | {s-$np_{sing}$} |
| #7 | search for entry with head np, num=sing | *die* 'that', cat=np/n, where n has num=sing | - | - |
| #8 | reduce arg np1 in #5 with head np in #7, dominated by s; shift arg n, num=sing | *die _ ontdekt diepe betekenissen* 'that _ discovers deep meanings' | { $s_{sing}$} | {s-$n_{sing}$} |
| #9 | search for entry with head n, num=sing | *Nederlander* 'Dutchman', cat=n, num=sing | - | - |
| #10 | reduce arg n in #7 with head n, dominated by s in #9 | *die Nederlander ontdekt diepe betekenissen* 'that Dutchman discovers deep meanings' | {$s_{sing}$} | {} |

Table 1: Generating a sentence

## 3. Methods

We will describe methods that store linguistic objects as objects in an OO lexicon, and methods, some borrowed from the Relational Model, that retrieve these objects efficiently and, yet better, fast. The methods have been implemented in ISO Prolog. They should obey the Resource Principle, which is stated thus: "Deploy working memory when performance is the key factor, and deploy external memory when storage is the main aspect". This principle is a practical phrasing of the insights that working memory will never be large enough to hold our current and planned number of linguistic objects, and that working memory is needed for real computation tasks and of the facts that working memory is faster than external memory, and that external disk space is abundant, and cheap. We refrain from implementing interfaces to third-party products, e.g. the (semi-commercial) Berkeley DB library for external storage of terms (SICS 2008), because we prefer 'light', compatible, and manageable interfaces, that are portable to other platforms.

The sections 3.1, 3.2, and 3.3 describe the techniques of index tables, caching, hashing and compression. In later sections their use is described.

### 3.1 Direct access and index tables

The 'Edinburgh' Prolog standard offers sequential read and write access to files. Although linear access is still efficient in complexity terms, searching will take an unacceptable amount of time in worst case. In ISO Prolog, the concept of a file has been improved on, and been replaced by the concept of a 'stream'. A stream is a file with random access, that is each byte in the file can be located in constant time. An index table is a table that relates unique, simplex values to the positions in the context where these values can be found. In our case, this translates to a table per feature that relates each feature's value to all occurrences of objects that contain that feature with that value. An index table can be implemented by a collection of clauses of a 2-ary relation with the first argument as key. Many Prolog systems use 'first-argument indexing' to locate the correct clause, given its key, in constant time. This technique is not part of ISO Prolog, but it is regarded as an essential facility for interpreting Prolog programs efficiently. A fast lexicon for semantic generation hinges on the stream concept and indexing techniques.

### 3.2 Caching

Internal memory can be much faster accessed than external memory. Applications often need the same data again, and again. This had led to the development of cache memories. A disk cache is a storage mechanism in working memory that keeps the most recently read data plus the data of adjoining sectors in a buffer. As soon as the application asks for some data, the buffer is consulted first, saving access time to the hard disk. When the required data does not fit in the buffer, an extra disk access is necessary. To exploit disk caching, the data must be ordered in a way that corresponds to a relevant search criterion. Disk caching might be implemented at the application level, as advocated by Ceri et al.

(1989), or at the operating system level, or both. The former is contra the Resource Principle.

## 3.3 Hashing and compression

Hashing (Knuth 1998) is a method that converts an arbitrary, complex value to a simplex one—the 'hash' or 'key'—by applying a hash function to it. Typically, hashing converts to an integer, because it is the most economic data type, and most easy to use by a computer program. A hash value enables an index table to be used. Searching for an object in a collection of $n$ objects, given an unhashed value of some constraint, will take $O(n)$ comparison operations. With a hashed value and objects that are hashed on the appropriate constraint by applying one and the same hash function, searching an arbitrary object only takes $O(1)$ time, assuming that the index table can be directly accessed. This is easy to implement in Prolog systems that can do first-argument indexing. As each piece of data ever to be searched for is fixed, the lexicon is a collection of 'static search sets'. For such sets, a 'perfect hash function' (PHF) can be designed, that is a function that will never assign the same hash to different data structures. However, depending on the complexity of the data structure, the hash can exceed the range of the integer data type on some computer platforms. A 'near perfect hash function' takes the integer range into account, but allows for duplicates ('collisions'). Duplicates increase space and time complexity. A 64-bit platform extends the integer range with orders of magnitude, compared to 32-bit, enabling a PHF to be used, yielding zero collisions.

Number grouping is a compression technique that replaces a set of adjacent integers by a range, starting with the lowest number, and ending with the highest number. The space complexity for a range is constant, instead of linear for a set of numbers. The time complexity for determining the subset of two ranges is constant, while it is linear for intersecting ordered sets.

The techniques described sofar are used for creating and accessing the object lexicon (section 3.4) and for the index tables (sections 3.5 and 3.6). Section 3.7 discusses lexical retrieval in general, and gives some complexity results.

## 3.4 Creating and accessing the object lexicon

The object lexicon and access methods are is generated by an off-line process, which is not subjected to the Resource Principle. During generation, the objects are checked for well-formedness and validness and are assigned an ID for reference only, corresponding to the order they are generated in. The process of creating them is not described here. Once they are created, they are linearized into a series of bytes, and stored as persistent objects, formatted as standard Prolog terms.

A persistent object can be seen as a variable-length 'record', a concept from the Relational Model. The object's ID corresponds to the 'record number'. The physical address of the persistent object's first byte is kept in an external access table. Obviously, the object lexicon and the access table have the same large number of entries. Storing them externally agrees with the Resource

Principle. The access table has records (addresses) of fixed size. Consequently, it can be addressed by a function that maps an ID=$n$ to the physical address of the $n$'th entry in the access table. The access table, opened as a stream, gives direct access to this address. The information to be found there points to the physical location of the $n$'th object in the object lexicon, which, as a stream, has direct access. Thus, objects are retrieved by ID via an indirect addressing scheme in $O(1)$ time, at the cost of one extra disk access, one function in working memory, and one auxiliary access table in external memory. Prolog's read predicate loads the objects as native Prolog terms.

Efficient data structures are the basis for efficient algorithms. Therefore, the persistent objects need to be ordered by some ciriterion. A parser would benefit from an ordering on the phonological field for lexical look-up and tagging purposes and would exploit a disk cache by accessing all objects with the same surface form using a buffered read operation. A generator would take advantage of an ordering on the most frequently used deep constraint for handling the agenda. As the generator only picks one carefully selected object at a time, no physical ordering seems to be beneficial. However, physical ordering by some criterion corresponds to a very compact index table for that criterion, because each entry's set of ID's can be represented by a range. It turns out, that a physical ordering on syntactic type will prove helpful to the generator. This makes clear, that a lexicon that has to be deployed to both a parser and a generator needs to meet two sets of requirements. As the class of syntactic types is much smaller than the class of surface forms, a physical ordering on the former will yield less and bigger ranges, and consequently a more compact index table than an ordering on the latter, saving more memory, in line with the Resource Principle and speeding up intersection operations.

### 3.5 Creating and accessing index tables for concept, type, and phon

For semantic generation, we require maximum performance on the retrieval of linguistic objects, specified by constraints on the features for semantic concept, syntactic type, and word form. For each object, these features are stored in three auxiliary index tables in working memory. If an entry does not exist, it is created and added to the index table together with the object's ID. If it does already exist, only the ID is added, in order. Number grouping is applied to the sets of ID's as much as possible. Each combinatory type is hashed by a PHF into a unique string, instead of an integer, because, theoretically, types are unlimited in size, which may yield too many collisions. Each word form and each semantic concept is hashed to an integer by applying a near-PHF to their alphabetic letters. The hash value is kept as small as possible by taking into account the usage frequency of letters in Dutch (http://www.onzetaal.nl/advies/letterfreq.php). This method is a variant on Huffman (1952) coding. Provisions are made for handling collisions, which result from lexical ambiguities. In general, retrieving all ID's of objects that satisfy either the concept, type, or phon constraint takes $O(1)$ time, when first-argument indexing is applied by the Prolog system to the resp. index tables, and zero disk accesses. The index tables have a space complexity that is a linear function with a small factor—due to number grouping—of the number of objects.

The index table on the feature that is used for physically ordering the objects has a space complexity that is a linear function of the number of the feature's values. The index tables are kept in working memory for fast access, in accordance with the Resource Principle.

The following lines are extracted from the type index, encoded by clauses of the predicate yx/2. This predicate encodes which objects contain which type.

```
yx('RHa/Hc',[45591+1130]).
yx('R/HcHc',[46722+568]).
yx('N/HcHc',[47291+575]).
```

The string 'RHa/Hc' in the first argument of the first clause is the hash of type s\np/np, shared by the objects with ID's in the range 45591 up to 46722, encoded by 45591+1130 in the second argument. The string 'R/HcHc' is the hash for type s/{np,np}, and the objects with ID's in the range 46722 up to 47291 share this type. The lines demonstrates the economic encoding of ranges of object ID's. Prolog's first-argument indexing technique can be applied to the predicate yx/2, because the hashes are unique values. To find some object with type s\np/np, the type is hashed, yielding 'RHa/Hc'. The hash is looked up in the type index, yielding a range of object ID's, including #46212. This ID is mapped to the 46212th entry of the access table by an access function. This entry holds the value 44,960,845, being the physical address of object #46212 in the object lexicon. The object lexicon is opened as a stream at the location of the physical address and the object is loaded, in this case the object for 'ontdekt' *discovers*, shown in figure 1.

### 3.6 Creating and accessing a meta index table

Additionally, the semantic generator may select objects that are specified by constraints on other features. We don't demand maximum performance on queries of this kind. For each linguistic object, each value and full path, starting at the top of the graph, of each feature ever to be retrieved, and not being concept, type or phon, is stored in one auxiliary index table in working memory. This 'meta' index table spans more than one feature. The ID's of all objects that specify the same value for some feature are number-grouped and stored in an external meta data table. Additionally, the set of ID's of objects that do not specify any value for the feature is calculated and stored. The storage address per feature-value pair is kept in the meta index table. Retrieval of all ID's of objects that match one of these constraints takes $O(1)$ time, when first-argument indexing is applied by the Prolog system, and one extra disk access. The space complexity of the meta index table is a linear function of the number of unique feature-value combinations, occupying only a fraction of the working memory, consistent with the Resource Principle.

### 3.7 General retrieval

Retrieval of linguistic objects is performed by executing a search task, specified by one or more constraints. When the search task is a graph, typically an argument sub graph of some object involved in the generation process, it is flattened into a series of constraints. Each constraint is looked-up in an appropriate index table. We distinguish strict and liberal constraints, which demand objects to match the constraint explicitly, or allow objects that are unspecified for the constraint respectively. Liberal constraints follow from graphs that may be underspecified. When a strict constraint is a restriction on semantic concept, syntactic type, or word form, the set of ID's of the objects satisfying the constraint is found immediately in the resp. index tables. For other features, including liberal contraints, the set of ID's is found after issuing an extra disk access to the meta data table. In all cases, a constraint is replaced by a set of ID's, corresponding to objects that are consistent or not inconsistent with the constraint. The objects that satisfy all constraints are identified by ID's that result from intersecting all sets of ID's. As these sets are ordered, calculating their intersection is a linear function of the size of the biggest set. The function factor may be very small when ranges are involved. In general, however, the ranges get fragmented after a few intersection operations.

By applying directly accessible pre-compiled access and index tables, searching is reduced to looking-up with great performance, in contrast to Prolog's own depth-first backtracking search algorithm, which is inefficient. In summary: finding one object, given its ID, takes $O(1)$ time and one disk access. Finding the set of ID's of objects satisfying a constraint on concept, type or phon only, takes $O(1)$ time and zero disk accesses. Finding the set of ID's of objects satisfying another constraint takes $O(1)$ time and one disk access. Finding the set of ID's of objects that satisfy $n$ arbitrary constraints takes $O(n)$ time and at most $n$ disk accesses.

## 4. Conclusions and discussion

We showed the design and access of an external lexicon to be used by a semantic generator component of an NLP system in a Prolog context. The major design goal was to develop a fast lexicon, as searching and finding complete linguistic data is the major activity of the generator. It appeared that some properties of our data (variables, recursion) prevented them from being stored in the Relational Model, e.g. in a relational database. Furthermore, it turned out that this model is inherently unsuitable to demonstrate fast performance on complete linguistic data, because data is spread across a large number of tables. The Object-Oriented Model proved to be a natural environment for our linguistic data when regarded as objects. The nature of categorial grammar, being the basis of our objects, and the nature of objects, being distinct copies, derived from one or more classes, introduce an exponential space demand when storing them in an object-oriented way. It was felt that this is an inevitable consequence of the way Delilah was designed. The space issue was remedied by putting in a bigger storage device, which is cheap nowadays. While the storage of objects is redundant, it was shown that an object can still be efficiently retrieved using advanced index tables in working memory, by compression techniques, and by implementing the lexicon and retrieval functions in ISO Prolog, using Prolog's first-argument indexing technique, and ISO Prolog's stream I/O. In this respect, ISO Prolog surpasses

'Edinburgh' Prolog with an order of magnitude. Efficient runtime performance was achieved by executing an off-line compilation process of the lexical resources into efficient data structures.

Currently, the lexicon holds ca. 70K fully inflected entries, specified by ca. 1,500 lemmas and over 200 constructions, occupying ca. 63 MB of disk space. The sum of all external index and auxiliary tables is under 3 MB. In the near future we hope to incorporate the valency information of the Alpino lexicon (Bouma et al. 2001). The information in Cornetto (Vossen 2007), including the Dutch Wordnet (Vossen 1998), and the Reference Database for Dutch (Maks et al. 1999), on how to phrase 'relations' can be used to improve the semantic generator, while word sense-disambiguation information may be used to improve Delilah's parser.

An important research objective will be how the way of organizing access to the stored lexicon, as described here, can be maintained when the lexicon increases in orders of magnitude. Sooner or later, the index tables will burst from working memory. At the same time, we have to investigate to what extent this functional organization of the lexicon, i.e. using pre-compiled indexes, reflects or should reflect cognitive insights. We feel that this question did not rise before we touched on the data ourselves, instead of applying standard techniques or third-party products.